\documentclass{article}
\usepackage{spconf,amsmath,amssymb,graphicx}
\title{Integrating Dependency Tree Into Self-attention for Sentence Representation}
\name{Junhua Ma$^{\star}$ \qquad Jiajun Li$^{\dagger}$ \qquad Yuxuan Liu$^{\star}$  \qquad Shangbo Zhou$^{\star}$  \qquad Xue Li$^{\dagger}$ }

\address{$^{\star}$ College of Computer Science,	Chongqing University, Chongqing, China \\
	$^{\dagger}$School of Information
		Technology \& Electrical Engineering, \\
		The University of Queensland, St Lucia, Qld, Australia\\
}
\begin{document}
%\ninept
%
\maketitle
\begin{abstract}
Recent progress on parse tree encoder for sentence representation learning is notable. However, these works mainly encode tree structures recursively, which is not conducive to parallelization. On the other hand, these works rarely take into account the labels of arcs in dependency trees. To address the both issues, we propose Dependency-Transformer, which applies a relation-attention mechanism that works in concert with the self-attention mechanism. This mechanism aims to encode the dependency and the spatial positional relations between nodes in the dependency tree of sentences. By an attention score way, we successfully inject the syntax information without affecting Transformer's parallelizability. Our model outperforms or is comparable to the state-of-the-art methods on four tasks for sentence representation and has obvious advantages in computational efficiency.

\end{abstract}

\section{Introduction}
\label{sec:intro}

In recent years, distributed sentence representations have been used in many natural language processing(NLP) tasks. 
Different composition operators have been used to map lexical representations to single sentence representations, such as recurrent neural network(RNN)\cite{conneau2017supervised,lin2017structured,liu2016learning}, convolutional Neural Networks(CNN)\cite{kalchbrenner2014convolutional,kim5882convolutionalneuralnetworksforsentence}, recursive convolutional neural network(RCNN)\cite{cho2014properties,zhao2015self} and Transformer\cite{vaswani2017attention}. Despite their success, these methods fail to explicitly take advantage of syntactic information in sentences.

%In contrast to earlier works\cite{tai2015improved,ahmed2019you,socher2011parsing,socher2013recursive,socher2014grounded,liu2017dynamic,xu2021treelstm}  that encoded sentences with sequence structure, tree-based approaches encode tree structure input recursively by using various composition functions. 
\begin{figure}[htbp]
	\centering\includegraphics[height=7cm]{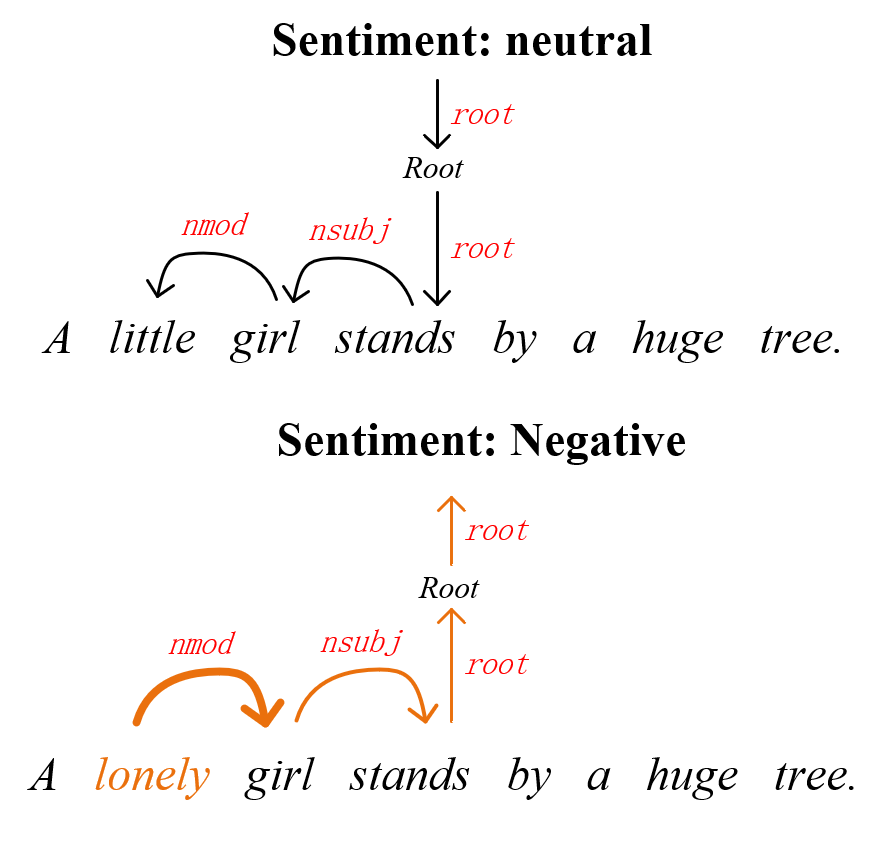}
	\caption{Adjust dependency intensity according to words' semantics.} \label{fig1}
\end{figure}
A parse tree shows the sentence's syntactic structure and contains rich grammatical information. To make use of the information, three recursive\cite{socher2011parsing,socher2013recursive,socher2014grounded} models were proposed to encode a sentence along with its parse tree, using several compositional functions to integrate tree nodes following a bottom-up manner. \cite{tai2015improved} proposed two LSTM-based models to encode sentences along with constituency tree and dependency tree recursively. Unfortunately, models based on RNN have a bad parallelizability. The heterogeneity of tree structure inputs makes it impossible to train multiple samples simultaneously in a batch, which further limits the parallelizability of the model.

On the other hand, Transformer has been extremely popular since its introduction because of its superior parallelism and performance. Tree-Transformer\cite{ahmed2019you} recursively encode words with their parents and children in the parse tree by a Transformer module. Although applied the transformer structure, the model still followed a recursive mechanism, which cannot be trained in parallel. Besides, dependency tree structure includes two parts of information: topological structure and types of dependency relations(label of dependency arc). These labels also provide critical information about the dependency tree, but few models can distinguish them well. This raises a question: can we utilize the information included in the syntax tree as much as possible without affect the original mechanism so that keep its parallelizability?  
\begin{figure*}[t]
	\centering\includegraphics[height=7cm]{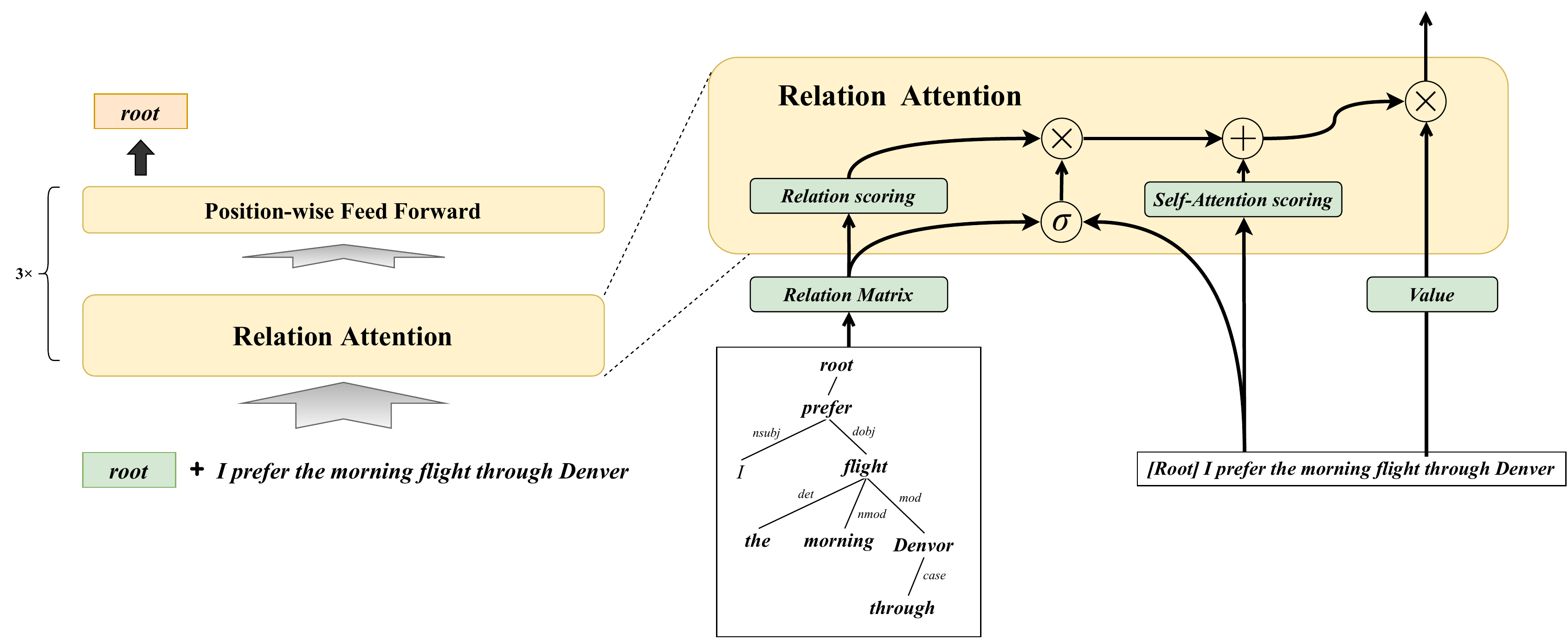}
	\caption{Overall structure of Dependency-Transformer} \label{fig1}
\end{figure*}

In this paper, we propose a Transformer-based model that applies the relation-attention mechanism, Dependency-Transformer, which provides a positive answer to the question above. The relation-attention mechanism integrates the dependency information into the self-attention mechanism in two steps. Firstly, unlike the recursive model, we tend to decompose the tree into a bunch of relations between words. We set learnable vectors for dependency relations and the relative positions in the tree, then pool them into scores and incorporate them into self-attention when the model computes attention scores for words pairs in parallel. Secondly, to link relation attention scores and the semantics of sentences shown in Figure 1, we implement a gating mechanism that enables these relation-attention scores to adjust the proportion that will be added to the self-attention scores by the lexical semantics.

We evaluate our model on four widely-used datasets of sentence representation tasks. Our model outperforms all baselines and has obvious advantages over SICK-E and SICK-R. For the SST-2 and MRPC, the model's performance is comparable to state-of-the-art methods. Moreover, our model improves the performance while achieving 6-8 times the training or testing speed of the recursive model. Furthermore, we conduct an ablation study and case study, which further validate the efficiency and rationality of the proposed mechanism. 

%Here we list our main contributions as following:
%\begin{itemize}
%	\item We propose Dependency-Transformer and relation-attention mechanism, retains Transformer's superior parallelizability while make better use of dependency tree.
%	\item We evaluate our model on four sentence representation tasks: SICK-E, SICK-R, SST-2 and MR. As results, we reach the state-of-the-art on two of them and are competitive on the others An ablation study and a case study are conducted to provide evidences for our work's effectiveness.
%\end{itemize}

\section{Proposed model}
Our model’s structure, shown in Figure 2, is based on a three-layer Transformer encoder. Given an input sequence $X = [{x_1},{x_2},...,{x_n}] \in {\mathbb{R}^{l \times d}}$ , a $[root]$ token will be inserted at the beginning of the sentence, which represents the corresponding $ROOT$ node in the parse tree, and then $X = [root,{x_1},{x_2},...,{x_n}] $. At the end, the model yield the sequence’s output $Y' = [{y_{root}},{y_0},{y_1},..,{y_n}]\in {\mathbb{R}^{(l+1) \times d}}$. We take the $ROOT$ node's output as the representation of the sentence.
%(some tasks will use the entire sentence output). 

The input of standard Transformer is the sum of two parts: the word embeddings and the positional vectors, which provide the distributional information about words of the training corpus and the words' positional information in the sentence. To enrich the expression of words, we introduce the level embeddings, which are defined as the levels of words in the parse tree (the distance from the root node). Similar to the positional vectors, we set learnable vectors for different words' levels and add them to the above embeddings together as the input of our model.

\begin{figure*}[htb]
	\centering\includegraphics[height=5.5cm]{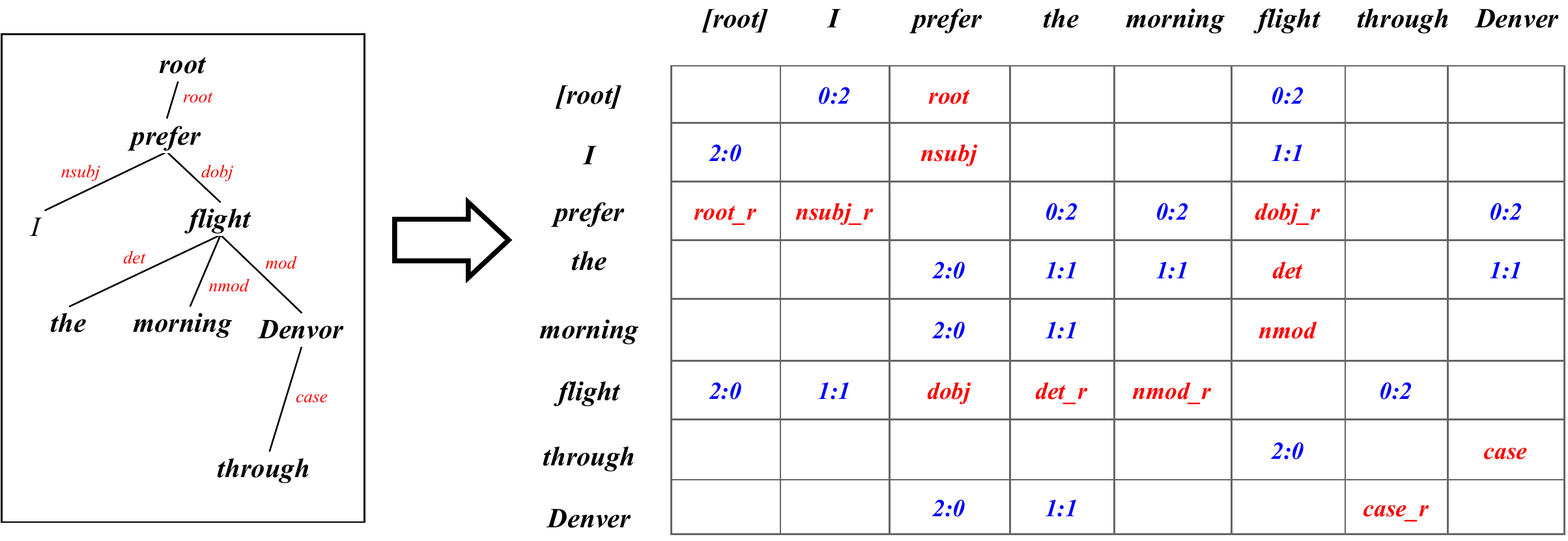}
	\caption{An example of relation matrix' constitution. Each element in the matrix represents the relation between two words. For the adjacent nodes, we assign the relations to the arc labels in the dependency tree(red color). For the non-adjacent nodes and the sum of their distances is less than a certain threshold(2 in the example), we concatenate their distances to the nearest common ancestor to represent the relations(blue color). } \label{fig2}
\end{figure*}
\vspace{-0.2cm}
\subsection{Dependency relation}

In self-attention of Transformer, attention scores are scaled-dot products of the hidden states of words, which may represent a kind of correlation between words. They are expected to capture the lexically semantic correlations. Therefore, we can also view the words' relation in the dependency parse tree as a correlation between words and inject it into self-attention as part of the attention scores. Correspondingly, These scores are expected to capture the syntactic information of the sentence.

As mentioned earlier, one of the previous works' limitations is the lack of utilization of relation types in the dependency tree, determining that our primary concern is how to incorporate these relations into Transformer and give them distinct attention. So in the implementation, we set a learnable vector for each types of relation. Different linear layers are applied at each attention head to calculate scores for these vectors. And then, these scores will be added to the original attention scores as a bias. The relation-attention score($word_i$ attend $word_j$) in a specific attention head is computed as follow(we omit the superscript of layer number for simplicity):

\begin{equation}
	S_{ij}^r = {r_{ij}}{V_r} \; ,
\end{equation}
where $r_{ij} \in \mathbb{R}^{d_r}$ denote the relation vector of $word_i$ to $word_j$, ${V_{r}} \in {{\mathbb{R}^{{d_r} \times {1}}}}$ is a learnable vector in current attention head.

\subsection{Semantic gated mechanism}
Note that the score of relation-attention and self-attention are independent of each other in such an implementation. This suggests that these scores only depend on the tree, limiting the model's fitting ability. Thus, we expect this score to be semantically correlated. A simple example of two sentences shows in Figure 1. In a sentiment analysis task, intuitively, even \emph{\textbf{little girl}} and \emph{\textbf{lonely girl}} in Figure 1 have the same type of dependency relation, the score of \emph{\textbf{girl}} attend to \emph{\textbf{little}} should be lower than \emph{\textbf{girl}} attend to \emph{\textbf{lonely}} because \emph{\textbf{lonely}} expresses a stronger sentiment. In other words, the mechanism needs to adjust the relation scores according to the semantics of the words, preventing these scores from affecting the self-attention improperly. we design a gating mechanism to link these scores to the dependent's semantics. That is, a gated score calculated by relation and query word determines the proportion of the relation-attention score that will be added to the attention score. We pool the relation embeddings and the Key word's hidden states to the gated scores and apply a tanh activation to scale them to $g \in [-1,1]$. We reformulate the attention score as follows:

\begin{equation}
	{S_{ij}} = (1-{g_{ij}} ) \odot S_{ij}^e + {g_{ij}} \odot S_{ij}^r \; ,
\end{equation}
\begin{equation}
	{g_{ij}} = sigmoid(({h_i}{W_{g,e}} + {r_{ij}}{W_{g,r}}){V_g}) \; ,
\end{equation}
where ${S_{ij}^e}$ denote the self-attention score of corresponding words. ${W_{g,r}} \in {{\mathbb{R}^{{d_r} \times {d_r} }}}$, ${W_{g,e}} \in {{\mathbb{R}^{{d_e} \times {d_r}}}}$, ${V_g} \in {{\mathbb{R}^{{d_r} \times 1}}}$ are trainable matrix and vectors which is independent in each attention heads. $h_i$ denote the hidden states of $word_i$, the hidden states here are the sum of word embeddings, position embeddings and level embeddings in the first layer: $h_k^0 = {e_k} + {p_k} + {l_k}$, are the output of last layer in the other layers. 
\subsection{Relation matrix infilling}
Nevertheless, if we only consider adjacent nodes in the tree structure, the relation matrix will be very sparse since every dependent only has one head, and the model benefits little from these tree structures. In order to enrich the input information, we fill the relation matrix by enrich the words dependency relation. Specifically, we identify non-adjacent word pairs whose distance is less than a certain threshold and view them as additional dependency relations. In the implementation, we use a relative position encoding of shortest paths\cite{yang2016position} to represent the relation between non-adjacent nodes. The specific construction of the relation matrix R(relation vector $r_{ij}$  corresponds to the element of the $i$-th row and $j$-th column of $R$) is as shown in Figure 2. 
%\subsection{Subtree attention mask}
%We apply subtree mask attention to half of the attention heads, which only allows query words to attend to their descendant in the dependency tree. This part of the attention heads can focus more on grammatical dependency by ignoring words that do not have grammatically directly or indirectly dependency. In addition, this subtree masking can simulate an accumulation follow the bottom-up manner. 
\begin{table}[t]
	
	\caption{Result of evaluation on four datasets.}
	
	\begin{center}
		
		\centering
		
		\renewcommand{\arraystretch}{1.3}
		
		\newcommand{\tabincell}[2]{\begin{tabular}{@{}#1@{}}#2\end{tabular}} 
		
		\setlength{\tabcolsep}{0mm}{%7可随机设置，调整到适合自己的大小为止
			
			\begin{tabular}{lcccc}
				
				\hline
				
				{\bf Models} & \tabincell{c}{\bf SICK-R\\{(MSE)}} &\tabincell{c} {\bf SICK-E\\(Acc.)(\%)} & \tabincell{c}{\bf SST-2\\(Acc.)(\%)} & \tabincell{c}{\bf MRPC\\(Acc.)(\%)} \\
				
				\hline

				LSTM\cite{tai2015improved} &     .2831 &       76.80 &       84.90 &       71.70 \\

				BiLSTM\cite{tai2015improved} &     .2736 &      82.11 &       87.50 &       72.70 \\
				
				\hline
				
				%$MV-RNN$\cite{so &      - &       75.5 &       82.9 &      66.91 \\

				RNTN\cite{socher2013recursive} &      - &       59.42 &       85.40 &      66.91 \\

				DT-RNN\cite{socher2013recursive} &     .3848 &      63.38 &       86.60 &      67.51 \\

				Tree-LSTM$_{DT}$\cite{tai2015improved} &     .2734 &         82.00 &       85.70 &  72.07 \\

				Tree-LSTM$_{CT}$\cite{tai2015improved} &     .2532 &      83.11 &         88.00 &      70.07 \\

				DC-TreeLSTM\cite{liu2017dynamic} &          - &       82.30 &       87.80 &          - \\

				BiTree-LSTM\cite{teng2017head} &     .2736 &          - &       90.30 &          - \\

				TagHyperTreeLSTM\cite{xu2021treelstm} &          - &       83.90 & {\bf 91.20} &          - \\
				
				\hline
				
				USE\cite{cer2018universal} & - & 81.15 &  85.38& {\bf74.96} \\
				
				Tree-Transformer$_{DT}$\cite{ahmed2019you} &     .2774 &      82.95 &      83.12 &      70.34 \\

				Tree-Transformer$_{CT}$\cite{ahmed2019you} &     .3012 &      82.72 &      86.66 &      71.73 \\

				\hline
				
				Ours& {\bf .2428} & {\bf 85.10} &       88.70 &      73.58 \\
				
				\hline
				
			\end{tabular}  
			
		}
		
		\label{tab1}
		
	\end{center}
\vspace{-0.7cm}
\end{table}
\section{Experiments}

\subsection{Experiment Setup}
%In this section, we introduce the datasets and the experiment setting in 5.1 and 5.2 respectively. We analyze the results of the evaluation in 5.3. In 5.4 and 5.5, we conduct an ablation study to further analyze our model.
\noindent\textbf{Dataset }We evaluate the proposed model on three typical sentence representations tasks: text classification, text semantic matching and paraphrase detection.
Text classification: Stanford Sentiment Treebank\cite{socher2011parsing} for sentiment analysis in binary(SST-2). 
% is consist of 9613 sentences in a 6920/872/1821 split, which are scraped from movie review. Each sentence is annotated with negative or positive.
Text semantic matching: The Sentences Involving Compositional Knowledge dataset(SICK)\cite{marelli2014semeval} contains two tasks: relatedness task (SICK-R) and the entailment task (SICK-E).
%is consist of 9927 sentence pairs in a 4500/500/4927 train/dev/test split, which are derived from existing image and video description datasets. The SICK dataset contain two tasks, For the relatedness task (SICK-R), each sentence pair is annotated with a relatedness score $y_r\in\left[1,\ 5\right]$. For the entailment task (SICK-E), each sentence pair has a class label $y_e\in\left\{0,\ 1,\ 2\right\}$, and corresponding relation between sentence pair is entailment/contradiction/neutral.
Paraphrase detection: Microsoft Research Paraphrase Corpus(MRPC)\cite{dolan2004unsupervised}.

\noindent\textbf{Baselines} We compare our model with several state-of-the-art models, including four basic architectures: LSTM and Bi-LSTM\cite{tai2015improved} , and eight tree-based models: DT-RNN\cite{socher2013recursive},  Tree-LSTM\cite{tai2015improved}, Bi-treeLSTM\cite{teng2017head} ,  DC-treeLSTM\cite{liu2017dynamic}, TagHyperTreeLSTM\cite{ha2016hypernetworks}, and two Transformer-based model: USE\cite{cer2018universal} and Tree-Transformer\cite{ahmed2019you}. Bi-treeLSTM is a encoder added bidirectional flow to tree-LSTM. DC-treeLSTM and TagHyperTreeLSTM took advantage of hypernetworks to generate dynamic parameters for different inputs. USE is a pretrained sentence encoder based on Transformer. For Tree-LSTM and Tree-Transfomer, they include two version which based on dependency tree or constituency tree respectively, and we distinguish them with subscripts CT and DT.

\noindent\textbf{Setting} For all tasks, we use the Stanford dependency parser\cite{manning2014stanford} to parse every sentence in datasets. Word embeddings are initialized with the 300-dimensional GloVe word vectors\cite{pennington2014glove}, and their weights will update during training. Our model consists of three layers, each of which uses a self-attention and relation-attention mechanism with six heads. We set the dimension of the relation vector to 30 and set the size of hidden states and Feed-Forward layers to 300. We take the output of $[root]$ token as the representation of this sentence. Our models are trained using AdaGrad\cite{duchi2011adaptive}  with a learning rate of 1e-3 and a batch size of 32. We report mean scores over five runs and trained our model on an Quadro RTX 5000 GPU, and used PyTorch 1.4.0 for the implementation.

For the text semantic matching task and paraphrase detection task, we use the same relatedness head module as tree-LSTM\cite{tai2015improved}, which computes the final output of sentence pair by both the distance and angle of the two sentence's representation. For the classification task, including SICK-E, SST and MRPC, we use cross-entropy loss and report accuracy. For SICK-R, we use a KL-divergence loss to measure the distance between the predicted and target distribution report Pearson correlation between prediction and gold label. The results are shown in Table 1.

\begin{table}
	
	\caption{Training and testing time of Dependency-Transformer and Tree-LSTM(Pytorch impletation) in one epoch on SICK-E. }
	
	\begin{center}
		
		\centering
		
		\renewcommand{\arraystretch}{1.3}
		
		\newcommand{\tabincell}[2]{\begin{tabular}{@{}#1@{}}#2\end{tabular}}  
		
		\setlength{\tabcolsep}{4mm}{%7可随机设置，调整到适合自己的大小为止
			
			
			\begin{tabular}{ccc}
				
				\hline
				
				Model & Testing & Training\\ 
				
				\hline

				Tree-LSTM  &         8s &       282s \\

				Ours&        \textless1s &        57s \\
				
				\hline
				
			\end{tabular}  
			
		}
		
		\label{tab2}
		
	\end{center}
\vspace{-0.7cm}
\end{table}
\subsection{Result}

For the SICK-E and SICK-R, our model is superior to all the baseline models and outperforms the top-performing models TreeLSTM and TagHyperTreeLSTM by an obvious margin: 0.0103 and 1.2\%. For the SST task, our model achieves comparable accuracy with TagHyper-TreeLSTM and Bi-Tree-LSTM and is better than the other tree-based models. Our model also obviously outperforms USE and Tree-Transformer, which are also based on Transformer. In the SST task, the model receives supervision at all the nodes in the tree, and the number of nodes in the dependency tree is less than the constituency tree\cite{tai2015improved}, which may limit the performance of our model on this dataset. For the MRPC task, our model is second only to USE and outperforms all the other tree-based models. 

We test the time cost of our model and tree-LSTM on SICK-E, shown in Tab 2. As can be seen from the table, our training and reasoning time is significantly less than Tree-LSTM, which indicates that parallelism improves the efficiency of our model significantly.

\subsection{Ablation Study}

We conduct an ablation study to validate the proposed mechanism. As shown in Tab 3, we compare the results of four models with different implementations. In the table, $Transfomer$ refers to the Transformer encoder with the same setting as our model. $DT_{lr}$ refers to Dependency-Transformer without level embedding and gating mechanism,  DT$_{rg}$ refers to Dependency-Transformer without the level embedding, $DT_{full}$ refers the full implementation of the Dependency-Transformer by adding gating mechanism to relation-attention. Clearly, the relation-attention mechanism enhances the standard Transformer's performance significantly. It shows that these types of relations all contribute to the tasks. And it also proves the effectiveness and rationality of the relation-attention and gating mechanisms. 

\begin{table}
	
	\caption{Performance of Dependency-Transformer with different implementation.}
	
	\begin{center}
		
		\centering
		
		\renewcommand{\arraystretch}{1.3}
		
		\newcommand{\tabincell}[2]{\begin{tabular}{@{}#1@{}}#2\end{tabular}}  
		
		\setlength{\tabcolsep}{0.5mm}{%7可随机设置，调整到适合自己的大小为止
			

			\begin{tabular}{lcccc}
				\hline
				{\bf Models} & \tabincell{c}{\bf SICK-R\\{(MSE)}} &\tabincell{c} {\bf SICK-E\\(Acc.)(\%)} & \tabincell{c}{\bf SST-2\\(Acc.)(\%)} & \tabincell{c}{\bf MRPC\\(Acc.)(\%)} \\
				
				\hline
				Transformer &            0.2833&       83.34 &            87.19&             72.41\\
				DT$_{lr}$ &            0.2545&       84.76 &            88.21&            73.01\\
				DT$_{rg}$ &            0.2526&       84.93 &            88.53&            73.11\\
				DT$_{full}$ &     0.2428 &       85.13 &       88.77 &       73.58\\
				\hline
			\end{tabular}  
		}
		
		\label{tab3}
	\end{center}
\vspace{-0.7cm}
\end{table}

\section{Conclusion}
We propose Dependency-Transformer, using a relation-attention mechanism that integrates dependency tree information into the self-attention mechanism. Additionally, we introduce a gating mechanism to link syntactic relation and semantic information. The model retains the parallel computing ability of the Transformer while enhanced by syntactic information. We show apparent advantages of the model in efficiency and performance on various sentence representation tasks compared to baselines.

\subsubsection*{Ackonwlegdement}
This work is supported by National Key R\&D Program of China (No.2017YFB1402405-5), and the Fundamental Research Funds for the Central Universities (No.2020CDCGJSJ0042). The authors thank all anonymous reviewers for their constructive comments.

%\section{REFERENCES}
%\label{sec:refs}

\end{document}